\documentclass[runningheads]{llncs}

\usepackage{eccv}

\usepackage{eccvabbrv}

\usepackage{graphicx}
\usepackage{booktabs}

\usepackage[accsupp]{axessibility}  % Improves PDF readability for those with disabilities.

\usepackage{mdframed}
\usepackage{xcolor}

\usepackage[pagebackref,breaklinks,colorlinks,citecolor=eccvblue]{hyperref}

\usepackage{orcidlink}

\usepackage[textsize=tiny]{todonotes}
\usepackage{multirow}

\begin{document}

\title{Agent3D-Zero:  An Agent for 
           Zero-shot 3D Understanding}

\titlerunning{Abbreviated paper title}

\author{Sha Zhang\inst{1, 2} \and
Di Huang\inst{3} \and
Jiajun Deng\inst{4}
\and Shixiang Tang\inst{3}
\and Wanli Ouyang\inst{2}
\and Tong He\inst{2}
\and Yanyong Zhang\inst{1}
}

\authorrunning{Sha.~Author et al.}

\institute{University of Science and Technology of China \and
Shanghai Artificial Intelligent Laboratory \and
The University of Sydney \and
The University of Adelaide, Australian Institute for Machine Learning}

\newcommand{\systemname}{\textbf{Agent3D-Zero}\xspace}
\newcommand{\lsystemname}{{Agent3D-Zero}\xspace}

\maketitle

\begin{abstract}
The ability to understand and reason the 3D real world is a crucial milestone towards artificial general intelligence. 
The current common practice is to finetune Large Language Models (LLMs) with 3D data and texts to enable 3D understanding. 
Despite their effectiveness, these approaches are inherently limited by the scale and diversity of the available 3D data.
Alternatively, in this work, we introduce \systemname, an innovative 3D-aware agent framework addressing the 3D scene understanding in a zero-shot manner. 
The essence of our approach centers on reconceptualizing the challenge of 3D scene perception as a process of understanding and synthesizing insights from multiple images, inspired by how our human beings attempt to understand 3D scenes.
By consolidating this idea, we propose a novel way to make use of a Large Visual Language Model (VLM) via \textit{actively} selecting and analyzing a series of viewpoints for 3D understanding.  
Specifically, given an input 3D scene, \systemname first processes a bird's-eye view image with custom-designed visual prompts, then iteratively chooses the next viewpoints to observe and summarize the underlying knowledge.
A distinctive advantage of \systemname is the introduction of novel visual prompts, which significantly unleash the VLMs' ability to identify the most informative viewpoints and thus facilitate observing 3D scenes. 
Extensive experiments demonstrate the effectiveness of the proposed framework in understanding diverse and previously unseen 3D environments.

\end{abstract}
\section{Introduction}
\begin{figure}[th]
  \centering
  \includegraphics[width=1.0\linewidth]{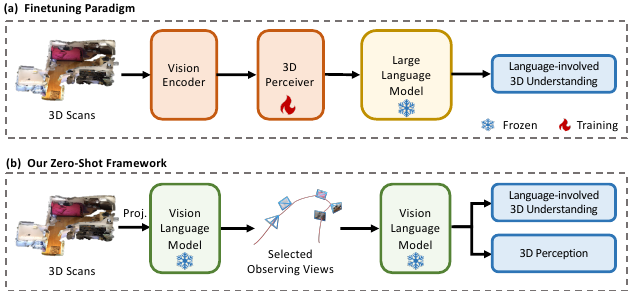}
  \caption{\textbf{Illustration of (a) finetuning-based paradigm and (b) our proposed zero-shot paradigm. } The finetuning-based paradigm exploits an external 3D perceiver, and finetunes it with a frozen LLM. On the contrary, our proposed zero-shot paradigm is simple and efficient, directly utilizing the VLM to actively select and interpret multiple observing views for zero-shot 3D task resolution.
  }
  \label{fig:intro}
\end{figure}

Understanding three-dimensional (3D) scenes~\cite{naseer2018survey} is a fundamental task in computer vision, especially vital for robotics~\cite{cadena2016robotic}, autonomous driving~\cite{zhang2024hvdistill, guo2021atsurvey}, and augmented reality applications~\cite{kalkofen2008ar}. 
% For example\dnote{Why give an example here?}, t
The emergence of an intelligent assistant who can fully comprehend the 3D world and find her way seamlessly in a given space is a crucial milestone on the path to artificial general intelligence.

Recently, the convergence of visual perception with Large Language Models (LLMs)~\cite{alayrac2022flamingo, liu2024llava} represents a significant leap forward, showcasing remarkable proficiency in a variety of 2D understanding tasks. Building on this advancement, extending these capabilities to the 3D realm seems a natural progression. This involves integrating 3D data into LLMs or Large Vision Language Models (VLMs) through fine-tuning~\cite{hong20243dllm, li20243dmit} and enabling the models to process 3D data formats directly. 
By leveraging their extensive prior knowledge, these LLMs/VLMs are poised to significantly enhance 3D understanding in open-world scenarios~\cite{azuma2022scanqa}.

However, collecting a large quantity of 3D data is a rather challenging task, which requires specialized equipment like depth cameras~\cite{horaud2016depthcamera} or LiDAR sensors~\cite{lemmens2007lidar}, along with sophisticated reconstruction algorithms~\cite{curless1996reconstruct}. Annotating the 3D data with textual descriptions can be much more difficult and labor-intensive compared to the 2D counterpart. Moreover, the diversity of publicly available 3D data is severely limited, with existing datasets often confined to CAD models~\cite{wu2015modelnet, hua2017objectNN}, indoor environments~\cite{dai2017scannet}, and autonomous driving scenarios~\cite{geiger2015kitti, caesar2020nuscenes}. Such issues motivate us to rethink the potential solutions toward 3D understanding with large foundation models.

Formally, in this work, we explore an alternative approach for developing a 3D-aware intelligent assistant. Contrary to the common practice of fine-tuning LLMs/VLMs on 3D data and text pairs, we introduce \systemname, an agent framework for VLMs addressing the 3D scene understanding in a zero-shot manner. Our approach draws inspiration from the human cognitive ability to comprehend the real world without the need for explicit 3D reconstruction, but observe the 3D scenario from multiviews. Humans, for example, can intuitively understand the spatial relationships among objects through visual perception, even without precise measurements of distance. 

By consolidating this idea, \systemname utilizes multiple images from diverse viewpoints, enabling VLMs, such as GPT-4V~\cite{openai2023gpt4}, to perform robust reasoning and achieve a reliable understanding of spatial relationships between objects. This approach allows \systemname to perceive the 3D world using the extensive knowledge embedded in pre-trained VLMs, thereby achieving zero-shot scene understanding.

Furthermore, \systemname is designed to actively select subsequent viewpoints for observing and reasoning about spatial information across multiple images.
While introducing multiple viewpoints enriches scene comprehension, it simultaneously imposes substantial memory and processing demands on current VLMs.
To address this, we propose a novel visual prompting technique termed \textit{Set-of-Line Prompting (SoLP)}. 
By employing bird's-eye view images, we delineate the scene's boundaries and superimpose a Cartesian coordinate system equipped with uniform grid lines and directional markers.
This straightforward yet effective strategy significantly enhances the VLM's capability to understand 3D spatial concepts.

We demonstrate the effectiveness of \systemname in various 3D reasoning and perception tasks.
In the 3D Question Answering task, our approach surpasses all related works in the ScanQA dataset, even without the use of annotated data for training or fine-tuning. 
Additionally, as a zero-shot method, \systemname gets advantages over the previous fine-tuning method in task decomposition and 3D-assisted dialog tasks.
Our contributions can be summarized as follows:
\begin{itemize}
\item We pioneer the design of \systemname, a 3D-aware Agent for scene understanding, which excels in zero-shot learning using only images, thereby eliminating the dependence on explicit 3D data structures such as point clouds or meshes.
\item We develop a comprehensive framework for proactive perception in agents, enabling the VLM to identify location and direction through inherent reasoning capabilities.
\item \systemname demonstrates exceptional performance across a range of tasks and multitasking scenarios, outperforming existing methodologies in experiments involving the ScanQA dataset, 3D-assisted dialogue, and zero-shot 3D segmentation.
\end{itemize}
\section{Related Work}

LLMs have shown remarkable potential in advancing artificial intelligence, demonstrating their effectiveness in various assessments~\cite{ouyang2022InstructGPT, chung2022flant5, chowdhery2023palm, zhang2023llama}. Initially limited to processing textual information, there has been a significant shift towards Multi-modal Large Language Models (MLLMs) to overcome these constraints by integrating multi-modal inputs.  This discussion outlines the transition from 2D to 3D scene understanding within MLLMs, providing a foundation for exploring advancements in multimodal learning.

\subsection{MLLM for 2D Scene Understanding}
The application of MLLMs for 2D scene understanding ~\cite{alayrac2022flamingo, chen2023minigpt, chen2023videollm, gong2023multimodalgpt, kim2021vilt, radford2021clip, liu2024llava, openai2023gpt4, kim2021vilt, li2022blip, li2023blip2} forms a critical foundation for our method, which utilizes VLMs to comprehend the 3D world in a zero-shot manner. As a precursor to our approach, insights into the operational mechanisms of VLMs offer valuable lessons. To establish a foundational model capable of interpreting 2D vision, researchers have pursued two distinct paradigms: training from scratch using internet-scale text-image pairs, exemplified by models such as CLIP~\cite{radford2021clip} by OpenAI, BLIP series ~\cite{li2022blip, li2023blip2} by Facebook AI, and ViLT~\cite{kim2021vilt} by Google; and fine-tuning existing LLMs with additional 2D vision data, as seen in models like GPT-4V~\cite{openai2023gpt4} and LLava~\cite{liu2024llava}. The former approach fosters a deeper, more intrinsic understanding of visual-textual relationships, whereas the latter is more cost-effective and often yields superior general performance. Leveraging these insights from LLMs and VLMs, the research community has extensively explored various applications and achievements of VLMs in 2D scene understanding, including image captioning~\cite{li2022mplug, li2022blip, li2023blip2}, visual question answering~\cite{antol2015vqa, schwenk2022okvqa}, semantic segmentation~\cite{yang2023som, rasheed2023glamm}, and zero-shot classification tasks~\cite{wu2023gpt4vis, naeem2023i2mvformer}.

\subsection{MLLM for 3D Scene Understanding}
Extending LLMs to 3D scene understanding marks a significant shift beyond traditional 2D analyses, embracing the complexity of our three-dimensional environment. Initial efforts focused on recognizing single 3D objects~\cite{wu2023gpt4vis}. Meanwhile, Models like Chat-3D~\cite{wang2023chat} and SpatialVLM~\cite{chen2024spatialvlm} enhance spatial understanding in a single image through innovative techniques. Recent advancements~\cite{hong20243dllm, li20243dmit, wang2023chat, chen2024spatialvlm, jatavallabhula2023conceptfusion} have adapted LLMs and VLMs for more comprehensive 3D spatial comprehension. 3D-LLM~\cite{hong20243dllm} pioneers the incorporation of 3D worlds into LLMs by training perceivers to interpret reconstructed 3D features. Similarly, 3DMIT~\cite{li20243dmit} focuses on refining the integration of 3D spatial data into LLMs through dedicated scene and object projectors. However, these approaches are limited by the unavailability of 3D scene data, restricting the scope of 3D understanding compared to the extensive resources available for 2D analysis.
Our method diverges by using VLMs to perform 3D tasks in a zero-shot manner, getting rid of the labor-intensive 3D data collection and giving an elegant receipt of interpreting and understanding complicated 3D scenes with actively selected 2D inputs.

\section{Method}
\begin{figure*}[t]
  \centering
  \includegraphics[width=\linewidth]{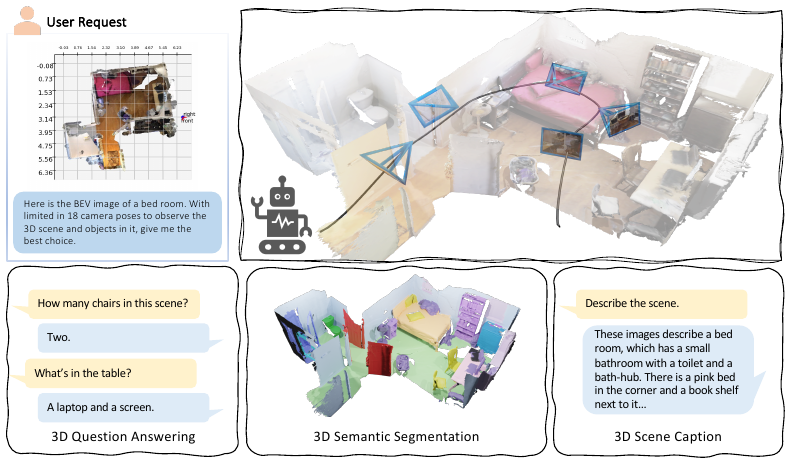}
  \caption{\textbf{System Overview of \lsystemname.} The upper segment illustrates our viewpoints-selection progress. We initiate the process by overlaying grid lines and tick marks on the Bird's Eye View (BEV) images, constituting the prompt along with a scene type description. This prompt guides the Vision Language Model (VLM) to retrieve camera poses for images observing the 3D scene. The lower section demonstrates the versatility of \lsystemname, showcasing its proficiency in addressing various 3D reasoning and perception tasks through strategic prompting and tool utilization. }
  \label{fig:overview}
  
\end{figure*}

\subsection{Overview}

\lsystemname presents a novel agent framework that utilizes VLMs for 3D scene understanding in a zero-shot manner. 
An overview is illustrated in Figure~\ref{fig:overview}. 
The process initiates with a bird's-eye view (BEV) image $I_b$ derived from a given 3D mesh $M$. 
The selection of $I_b$ serves a critical purpose: it provides a comprehensive layout of the scene, which is instrumental in planning the camera viewpoints that are crucial for a thorough scene analysis.
Given $I_b$ and the camera intrinsic metric $K$, we aim to strategically plan $N$ camera viewpoints. 
These are defined by the extrinsic matrix $\mathcal{T} =\{(R, t)_i | i \in [1,N] \}$, with $R$ denoting a 3x3 rotation matrix for the camera's orientation and $t$ a 3x1 translation vector for its position. This is formalized as:
\begin{equation}
\mathcal{T} = \textbf{VLM}(I_b, P_b) \label{eq:1},
\end{equation}
where the $P$ represents the textural prompt designed to guide the VLM. The prompt, structured to communicate the goal and specific requirements, might look as follows:

\begin{mdframed}[backgroundcolor=gray!20]
\color{gray}{\% Goal}\color{black} \\
Given a bird's-eye view of a scene, please provide N pictures to comprehensively understand the scene... \\
\color{gray}{\% Specific question}\color{black} \\
Could you suggest camera positions and orientations for each shot?

\end{mdframed}

Upon determining the camera extrinsic matrix $\mathcal{T}$, new images $\mathcal{I} =\{I_i | i \in [1,N] \}$ are rendered from the 3D mesh:
\begin{equation}
    I_i = \pi(R_i, t_i, K, M),
\end{equation}
where $\pi$ represents the rendering process. 
Subsequently, VLMs analyze images from these selected viewpoints, synthesizing insights for a coherent scene interpretation.
This empowers the agent with a detailed understanding of the scene, enabling it to tackle diverse scene-understanding tasks. 
With a task-specific prompt $P_t$, \lsystemname can query the VLM to address specific questions:
\begin{equation}
    A_t = \textbf{VLM}(\mathcal{I}, P_t), \label{eq: A_t}
\end{equation}
where $A$ symbolizes the answer to the question. 

In practice, instead of directly asking the model to output $N$ camera positions, we make the VLM iteratively output $N^\prime$ each time, enabling the VLM to select the viewpoint with previous experiences.

Regrettably, when relying on raw BEV image inputs, the VLM struggles to generate meaningful viewpoints due to inherent limitations in distance measurement. To address this challenge, we propose a visual prompt, the Set-of-Line Prompting (SoLP) technique based on BEV images. Subsequent section~\ref{subsection:SoLP} elaborates on the concept and application of SoLP. This is followed by a detailed exploration of \lsystemname's capabilities in zero-shot 3D scene understanding, focusing on its reasoning (section~\ref{subsection:Reasoning}) and perception (section~\ref{subsection:perception}) skills.

\subsection{Set-of-Line Prompting}
\label{subsection:SoLP}

Recognizing the challenge of precise location determination faced by VLMs, we introduce SoLP as an innovative solution to improve the VLMs' understanding of mathematical and spatial concepts. Traditional prompt engineering focuses on textural inputs for LLMs. However, our approach aims to adopt similar methodologies for visual inputs, enhancing the VLMs' planning and localization capabilities. Given that VLMs primarily produce textual outputs, the auxiliary visual prompts added to input images must be both interpretable by the models and describable in textual form.

Drawing inspiration from how humans interpret maps, where longitude and latitude are used to specify locations on a spherical surface and the Cartesian coordinate system represents points in a plane, we devised SoLP. This method involves superimposing grid lines and tick marks onto a BEV image, transforming the original image 
$I_b$ into a prompted image $I_b^p$.
Consequently, the process described in Equation \ref{eq:1} and \ref{eq: A_t} is refined as shown in Equation \ref{eq:2} and \ref{eq: A_t2}:

\begin{equation}
    \mathcal{T} = \textbf{VLM}(I_b^p, P_b^p)  \label{eq:2},
\end{equation}

\begin{equation}
    A_t = \textbf{VLM}(\mathcal{I}, P_t^p), \label{eq: A_t2}
\end{equation}
where $P_t^p$ represents the text prompt paired with the proposed SoLP.

SoLP not only aids in enhancing the VLM's comprehension of the scene's geometric aspects but also enables the generation of more precise camera poses. While directly determining $\mathcal{T}$ poses a significant challenge for VLMs, the introduction of SoLP facilitates this process by incorporating additional output format controls. This allows for the specification of camera positions as grid points within the image (e.g., (0, 0)) and orientations from a predefined set of directions [`left', `right', `front', `back'], simplifying the VLM's task. 
Compared with $P_t$, $P_t^p$ has additional format control sentences:

\begin{mdframed}[backgroundcolor=gray!20]
\color{gray}{\% Output format control}\color{black} \\
The position can be present as the grid point in the picture, like (0, 0). The orientations can be chosen from [`left', `right', `front', `back']. 
\end{mdframed}

Upon acquiring the camera poses $\mathcal{T}$, we can render the corresponding images $\mathcal{I} =\{{I}_i | i \in [1,N] \}$ within the 3D scene, setting the stage for subsequent analyses and evaluations.

\subsection{Exploration of 3D Reasoning Capabilities}
\label{subsection:Reasoning}

Upon processing the scene images, \lsystemname empowers VLMs to assimilate comprehensive information about the scene, thereby equipping the model to tackle a wide array of downstream tasks, including question answering (QA), caption generation, and dialogues. As indicated by the framework outlined in Equation~\ref{eq: A_t}, \lsystemname adapts to diverse tasks by employing specific task-oriented prompts $P_t$. 
An illustrative prompt $P_t$ is provided below to demonstrate how \lsystemname can guide VLMs in understanding a 3D scene through images from varied viewpoints and subsequently respond to a series of queries:

\begin{mdframed}[backgroundcolor=gray!20]
\color{gray}{\% Task prompt}\color{black} \\
Understand a 3D scene without direct access to the point clouds but only images from different viewpoints. Later, I'll ask you a series of questions about the scene, and I'd like your responses one-by-one with correspondence number, in the order the questions are presented. Please keep each response short and clear. \\
\color{gray}{\% Example questions}\color{black} \\
Examples: questions: [1. How many chairs are around the table? 2. what's the color of the table? 3. Where is the beige wooden working table placed? 4. What is in the corner of the bath? ]. \\
\color{gray}{\% Expected responses}\color{black} \\
Answers: [1. 3 2. Brown 3. right of tall cabinet 4. shower]
\end{mdframed}

Similar to a 3D-LLM~\cite{hong20243dllm}, \lsystemname boasts the capability to execute various downstream tasks using a singular model framework. This is a departure from most prior approaches, which typically specialize in a single aspect of 3D reasoning. This highlights the general applicability and adaptability of \lsystemname, showcasing its potential to serve as a versatile tool for 3D scene analysis and understanding.

\subsection{Exploration of 3D Perception Capabilities}
\label{subsection:perception}

\lsystemname distinguishes itself not only through its proficiency in language-centric 3D tasks but also by adeptly handling conventional perception tasks, such as 3D semantic segmentation. Although \lsystemname does not possess inherent perception abilities, it functions as an agent that effectively utilizes a variety of vision tools to enhance its recognition capabilities. 
We define the tool function as $f$, the per-view perception results as $R_i$.
This process can be formalized as:
\begin{equation}
    R_i = \textbf{VLM}(I_i, P_f, f),
\end{equation}

Taking the 3D semantic segmentation task as an illustrative example, our approach unfolds in two primary steps. The first step involves performing 2D semantic segmentation on each image selected for analysis. Second, with the aid of depth information, the 2D segmented results are back-projected into 3D point clouds to construct a comprehensive 3D semantic map.

Inspired by the Set-of-Mark (SoM) method \cite{yang2023set}, \lsystemname utilizes a Segment Anything Model (SAM) to segment the imagery into distinct regions initially without semantic labels. This segmentation facilitates the arrangement of the regions within each image, which, in turn, aids the VLM in assigning accurate semantic labels to each region. By annotating every selected image with semantic labels and then employing backprojection and concatenation techniques, \lsystemname successfully accomplishes 3D semantic segmentation.%the task of 3D semantic segmentation.
\section{Experiment}

\subsection{Datasets and Metrics}
\label{subsection:dataset}

We conduct experiments on the ScanQA dataset~\cite{azuma2022scanqa} and Scannet v2 dataset~\cite{dai2017scannet} to evaluate the performance of 3DQA and semantic segmentation, respectively. Moreover, we follow the practice of 3D-LLM\cite{hong20243dllm} to separate the scene dataset to evaluate the other 3D reasoning tasks.

\noindent \textbf{ScanQA}~\cite{azuma2022scanqa}: 
This dataset comprises over 40,000 human-annotated question-and-answer pairs, with each pair grounded within the objects of 800 indoor 3D scenes from the ScanNet dataset. Our experiments are conducted exclusively on the evaluation and test subsets of this dataset.

\noindent \textbf{ScanNet v2}~\cite{dai2017scannet}:  
ScanNet v2 is an extensive collection of 1,513 indoor scenes, meticulously annotated and equipped with multi-view RGB-D images alongside reconstructed meshes. For the task of semantic segmentation, this dataset offers 20 distinct classes of annotated 3D objects.

\noindent \textbf{3D-LLM held-in dataset}~\cite{hong20243dllm}: To compare our method with the most related work (i.e. 3D-LLM~\cite{hong20243dllm}), we follow it to evaluate the performance in 3D-assisted dialogue and task decomposition. This dataset contains 300k 3D-language pairs conducted by the data-generation pipeline in 3D-LLM, which is based on Scannet~\cite{dai2017scannet}, Habitat-Matterport~\cite{ramakrishnan2021habitat}, and Objaverse\cite{deitke2023objaverse}.

\noindent \textbf{Evaulation metrics}:  
We employ a suite of metrics to quantitatively evaluate the performance on language-related 3D reasoning tasks. These include BLEU~\cite{papineni2002bleu}, ROUGE-L~\cite{lin2004rouge}, METEOR~\cite{banerjee2005meteor}, and CIDEr~\cite{vedantam2015cider}, which collectively assess the quality of generated textual responses. METEOR, ROUGE-L, and CIDEr are designed to capture meaning and semantic coherence, often rewarding answers that are thematically consistent and informative, even if they diverge in exact wording. On the other hand, BLEU and EM metrics focus heavily on n-gram precision, rewarding exact matches between the predicted and reference sequences. For 3D Semantic Segmentation, Mean IoU (MIoU) serves as our primary metric, offering a comprehensive measure of segmentation accuracy.

\begin{table}[t]
    \caption{\textbf{Performance comparison on the ScanQA validation set.} `Two-stage' means the models use explicit object representations. `Fine-tune' means extra training. Our proposed Agent3D-Zero is training-free.  B-1, B-4 denote BLEU-1, BLEU-4\cite{papineni2002bleu}respectively. Our model outperforms all related models and the baseline model for evaluation metrics METEOR\cite{banerjee2005meteor}, ROUGE-L\cite{lin2004rouge}, and CIDEr\cite{vedantam2015cider}.
    }
    \vspace{-5pt}
    \small
    \setlength\tabcolsep{6.2pt}
    \centering
    \resizebox{0.99\textwidth}{!}{%
    \begin{tabular}{l l|cccccc}
    \hline
        ~ & ~ & B-1 & B-4 & METEOR & ROUGE-L & CIDEr & EM \\ 
        \midrule[1pt]
        \multirow{3}{*}{Two-stage}&  VoteNet+MCAN & 28.0  & 6.2 & 11.4 & 29.8 & 54.7 & 17.3 \\ 
        ~ & ScanRefer+MCAN & 26.9  & 7.9 & 11.5 & 30 & 55.4 & 18.6 \\ 
        ~ & ScanQA & 30.2 & 10.1 & 13.1 & 33.3 & 64.9 & \textbf{21.0} \\ \hline
        \multirow{7}{*}{Fine-tune}& flamingo-SingleImage & 23.8  & 8.5 & 10.7 & 29.6 & 52 & 16.9 \\ 
        ~ & flamingo-MultiView & 25.6 & 8.4 & 11.3 & 31.1 & 55 & 18.8 \\ 
        ~ & BLIP2-flant5-SingleImage & 28.6 & 5.1 & 10.6 & 25.8 & 42.6 & 13.3 \\ 
        ~ & BLIP2-flant5-MultiView & 29.7 & 5.9 & 11.3 & 26.6 & 45.7 & 13.6 \\ %\hline
        ~ & 3D-LLM (flamingo)  & 30.3 & 7.2 & 12.2 & 32.3 & 59.2 & 20.4 \\ 
        ~ & 3D-LLM (BLIP2-opt) & 35.9 & 9.4 & 13.8 & 34.0 & 63.8 & 19.3 \\ 
        ~ &3D-LLM (BLIP2-flant5) & \textbf{39.3} & \textbf{12.0} & {14.5} & {35.7} & {69.4} & 20.5\\ \hline
        \multirow{3}{*}{Zero-Shot} & LLaVA-SingleImage  & 7.1 & 0.3 & 10.5 & 12.3 & 5.7 & 0.0\\
        ~ & {\lsystemname(random)} & {16.4} & {2.1} & {12.2} & {26.9} & {40.0} & {4.9}\\
        ~ & {\lsystemname(selected)} & {28.6} & {4.4} & \textbf{16.0} & \textbf{37.0} & \textbf{71.8} & {17.5}\\
        \hline
    \end{tabular}
    }
    \label{tab:scanqa-val}
    % \vspace{-5pt}
\end{table}

\begin{table}[t]
    \caption{\textbf{Performance comparison on the ScanQA test set.} B-1, B-4 denote BLEU-1, BLEU-4 respectively. Our model outperforms all related models and the baseline model for evaluation metrics METEOR, ROUGE-L, and CIDEr.}
    \centering
    \vspace{-5pt}
    \small
    \setlength\tabcolsep{6.2pt}
    \centering
    \resizebox{0.99\textwidth}{!}{%
    \begin{tabular}{l l|cccccc}
        \hline
        ~ &~ &  B-1 & B-4 & METEOR & ROUGE-L & CIDEr & EM \\ 
        % \hline
        \midrule[1pt]
        \multirow{4}{*}{Two-stage}& SingleImage+MCAN & 16.5 & 0.0 & 8.4 & 21.5 & 38.6 & 15.8 \\ 
        ~ &VoteNet+MCAN* & 29.5 & 6.0 & 12.0 & 30.9 & 58.2 & 19.7 \\ 
        ~ &ScanRefer+MCAN* & 27.9 & 7.5 & 11.9 & 30.7 & 57.4 & 20.6 \\ 
        ~ &ScanQA* & 31.6 & \textbf{12.0} & 13.5 & 34.3 & 67.3 & \textbf{23.5} \\ \hline
        \multirow{3}{*}{Fine-tune} &3D-LLM (flamingo) & 32.6 & 8.4 & 13.5 & 34.8 & 65.6 & 23.2 \\ 
        ~ &3D-LLM (BLIP2-opt) & 37.3 & 10.7 & 14.3 & 34.5 & 67.1 & 19.1 \\ 
        ~ &3D-LLM (BLIP2-flant5) & \textbf{38.3} & 11.6 & {14.9} & {35.3} & {69.6} & 19.1 \\ \hline
        Zero-Shot & {\lsystemname} & {31.4} & {5.1} & \textbf{16.9} & \textbf{39.3} & \textbf{77.5} & {21.3} \\
        \hline
    \end{tabular}
    }
    \label{tab:scanqa-test}
    % \vspace{-1em}
    \vspace{-5pt}
\end{table}

\subsection{Compared Methods}
To evaluate the effectiveness of \lsystemname, we compare it with existing methods across several dimensions: Two-stage, Finetune, and Zero-Shot types.

\noindent\textbf{Two-stage methods}: These approaches, including combinations like VoteNet\cite{ding2019votenet}+MCAN\cite{yu2019mcan}, ScanRefer\cite{chen2020scanrefer}+MCAN\cite{yu2019mcan}, and ScanQA\cite{azuma2022scanqa}, first recognize objects in the 3D scene, then integrate language information to address Scan Question Answering tasks.

\noindent\textbf{Finetune methods}: Following the fine-tuning paradigm\cite{hong20243dllm}, this category includes models like flamingo\cite{alayrac2022flamingo} and BLIP2-flant5\cite{li2023blip2, chung2022flant5}. They train a perceiver to adapt LLM/VLMs to comprehend reconstructed 3D representations through a three-step process: encoding in 2D/3D, fine-tuning perceivers to align features with VLMs, and producing language-based answers. The inputs for these perceivers vary, being single-image, multi-view, or reconstructed 3D features.

\noindent\textbf{Zero-Shot methods}: These methods, such as LLaVA\cite{liu2024llava} and our \lsystemname, utilize VLMs to understand 3D scenes without any additional training. In all experiments, we utilize GPT4-V~\cite{openai2023gpt4} as the VLM in \lsystemname.%Notably, \lsystemname enhances this approach by designing visual prompts to select viewpoints before inputting the corresponding images.

\subsection{Zero-shot Performance}
\label{subsection:performance}

\noindent{\textbf{ScanQA.}} 
In the 3D-QA task, models must utilize visual data from comprehensive RGB-D indoor scans to answer textual queries about the 3D scene. Unlike conventional 2D-QA, models face challenges in spatial understanding and object identification from textual descriptions in 3D contexts.

Evaluation results (Table~\ref{tab:scanqa-val}) reveal that, despite its zero-shot operation, \lsystemname demonstrates competitive performance against other methods. Notably, with iteratively selected image inputs, \lsystemname outperforms previous benchmarks with scores of 16.0 vs 14.5 in METEOR, 37.0 vs 35.7 in ROUGE-L, and 71.8 vs 69.4 in CIDEr. However, it does not lead in EM or BLEU metrics, which prioritize n-gram precision. This discrepancy highlights the limitation of these metrics in evaluating the nuanced understanding and flexibility of zero-shot methods. Metrics like METEOR, ROUGE-L, and CIDEr are better suited for appreciating diverse expressions and capturing the essence of responses, as they do not solely reward exact matches.

Furthermore, Table~\ref{tab:scanqa-test} highlights \lsystemname's superior performance in most metrics compared to previous methods (77.5 in CIDEr, 39.3 in ROUGE-L, and 16.8 in METEOR), indicating its robust capability in 3D QA.%scene understanding.

\begin{table*}[h]
\caption{\textbf{Performance comparison on the Held-In Dataset}, which is introduced in 3D-LLM~\cite{hong20243dllm}. Our zero-shot method outperforms related methods.}
\vspace{-5pt}
\setlength\tabcolsep{6.2pt}{
\resizebox{0.99\textwidth}{!}{%
\begin{tabular}{ll|cccc}
\hline
Tasks& Models   & BLEU-1  & BLEU-4  & METEOR & ROUGE-L\\
% \hline
\midrule[1pt]
\multirow{8}{*}{3D-assisted Dialog} & flant5   & 27.4    & 8.7    & 9.5    & 27.5   \\
 & flamingo-SingleImage & 29.4   & 9.4    & 10.0   & 26.8   \\
 & flamingo-MultiView   & 30.6   & 9.1    & 10.4   & 27.9   \\
 & BLIP2-flant5-SingleImage     & 28.4   & 9.1    & 10.2   & 27.4   \\
 & BLIP2-flant5-MultiView & 32.4   & 9.5    & 11.0   & 29.5   \\
 & 3D-LLM (flamingo)    & 35.0  & 10.6   & 16.0   & 34.2   \\
 & 3D-LLM (BLIP2-opt)   & \textbf{39.6} & 16.2 & 18.4   & 38.6 \\
 & 3D-LLM (BLIP2-flant5)    & 39.0   & \textbf{16.6}  & {18.9} & \textbf{39.3}
 \\
 & {\lsystemname(random)}     & {26.9}  & 
 {7.1}  & {17.2} & {30.9} \\
 & {\lsystemname(selected)}   & {32.8} & 
 {9.8}  & \textbf{19.3} & \textbf{39.3} \\
 \hline
\multirow{8}{*}{Task Decomposition} & flant5 & 25.5   & 6.0  & 13.9   & 28.4   \\
 & flamingo-SingleImage & 31.4  & 7.1   & 15.6    & 30.6  \\
 & flamingo-MultiView   & 33.1 & 7.3   & 16.1   & 33.2  \\
 & BLIP2-flant5-SingleImage     & 32.2  & 6.9    & 15.0   & 31.0\\
 & BLIP2-flant5-MultiView & 33.1 & 6.9   & 15.5   & 34.0\\
 & 3D-LLM (flamingo)    & 32.9  & 6.4    & 16.0   & 33.5   \\
 & 3D-LLM (BLIP2-opt)   & {34.1} & {7.6} & {16.5} & 35.4 \\
 & 3D-LLM (BLIP2-flant5)    & 33.9    & 7.4    & 15.9  & {37.8}  \\ 
 & {\lsystemname(random)}     & {33.8}  & 
 {6.7}  & {16.7} & {36.6} \\
 & {\lsystemname(selected)}   & \textbf{42.0}  & 
 \textbf{15.5}  & \textbf{22.9} & \textbf{45.1} \\
 \hline
\multirow{2}{*}{3D Captioning}& {\lsystemname(random)} & 26.1   & 1.0  & 13.9  & 14.3   \\
 & {\lsystemname(selected)}   & \textbf{29.5}  & 
 \textbf{7.2}  & \textbf{15.9} & \textbf{16.1} \\
 \cline{1-6}
\end{tabular}
}
}

\vspace{-5pt}
\label{tab:held-in}
\end{table*}

\noindent\textbf{3D-assisted dialog.}
3D-assisted dialog systems incorporate spatial awareness to enhance conversational interactions.
According to results (Table~\ref{tab:held-in}), \lsystemname matches or exceeds former methods in key metrics like ROUGE-L and METEOR, reaffirming its capability in nuanced conversational contexts. BLEU metrics are discussed in the previous ScanQA analysis.

\begin{figure}[h]
  \centering
  \includegraphics[width=0.9\linewidth]{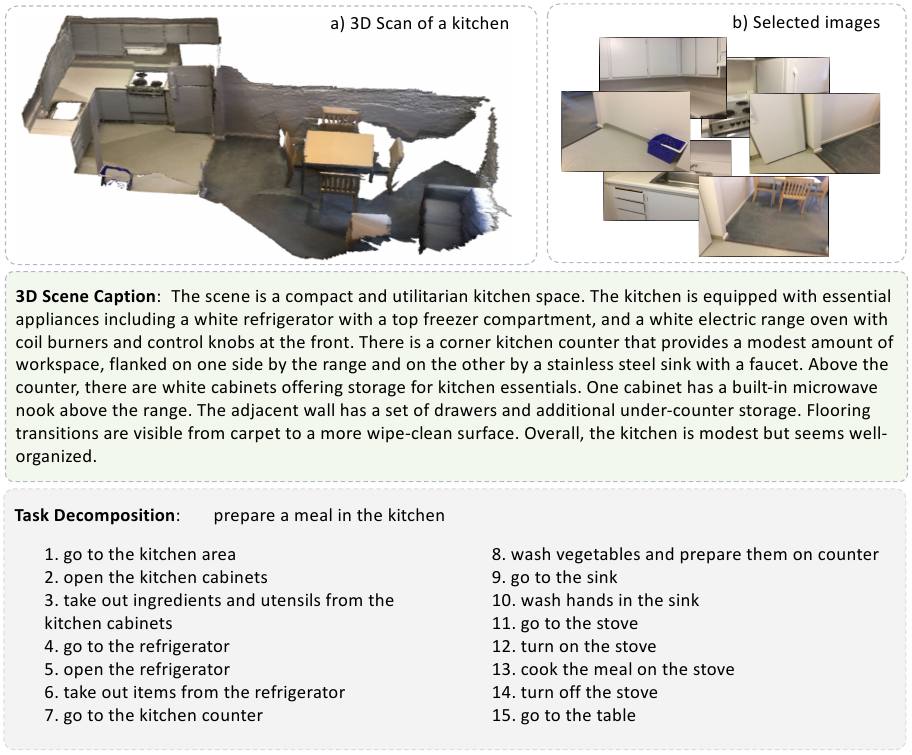}
  \caption{\textbf{Visualization of 3D Scene Caption and Task Decomposition of \lsystemname.} The top part presents the raw 3D scan and some of the images selected from different viewpoints.  We show examples of 3D Scene Caption and Task Decomposition at the bottom. }
  \label{fig:cap}
  \vspace{-5pt}
\end{figure}

\begin{figure}[h]
  \centering
  \includegraphics[width=0.8\linewidth]{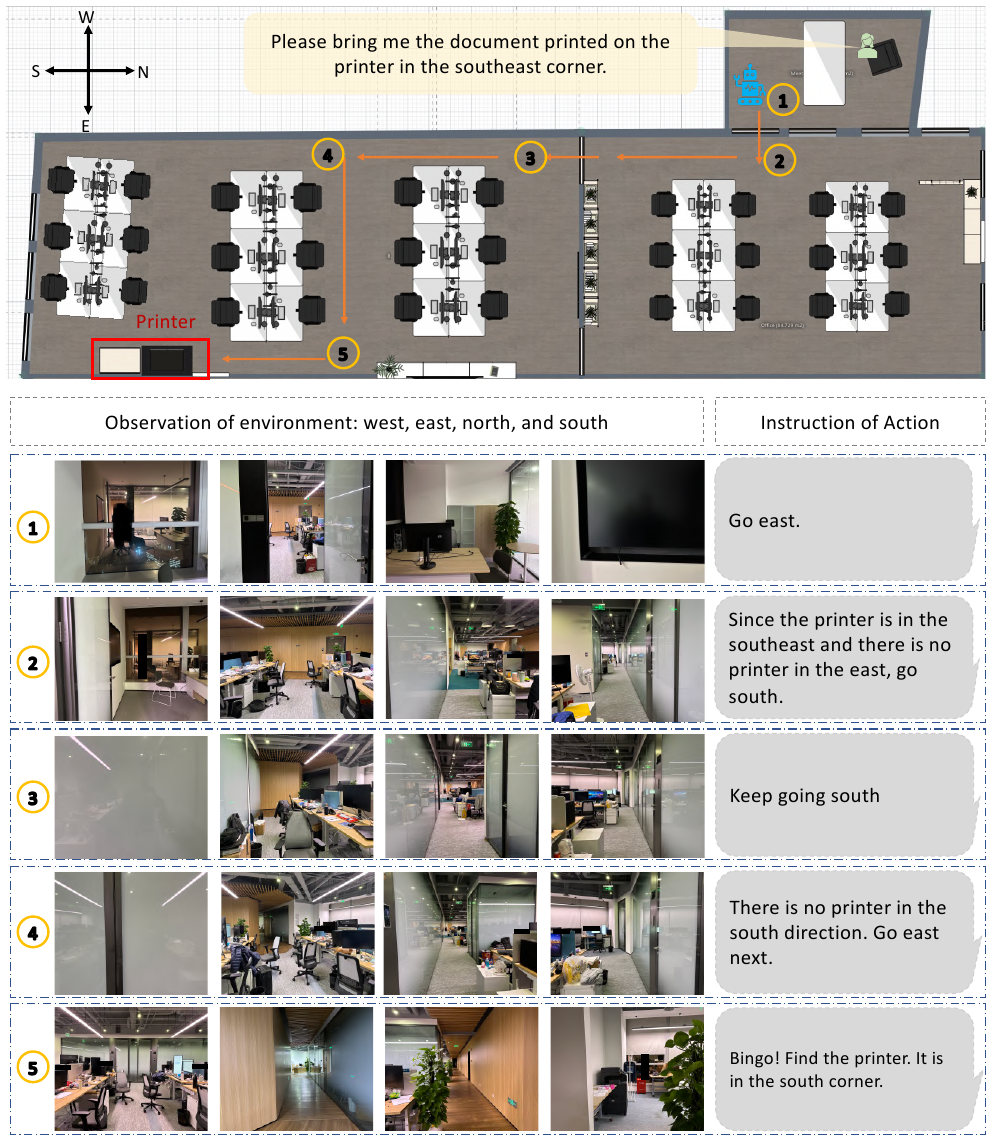}
  \caption{\textbf{Visualization of navigation in real world.} The top section introduces the navigation task and provides an overview of an office setting. Subsequent rows feature observations of the environment, with GPT-4v-generated instructions on the right. The visualization concludes with the VLM successfully locating the printer, thereby accomplishing the task in an unfamiliar environment.}
  \label{fig:demo}
  \vspace{-5pt}
\end{figure}

\noindent\textbf{3D scene caption.}
3D captioning requires a holistic scene understanding, extending beyond mere image analysis to comprehend spatial arrangements and object interactions. 
As the validation set for 3d scene captioning in the 3D-LLM held-in dataset only focuses on single 3d objects, we randomly select 51 scenes from the train set in the 3D-LLM held-in dataset to evaluate \lsystemname 's 3d scene caption ability.
The results are presented in Table~\ref{tab:held-in}. 
\systemname's performance is comparable to the baseline method, indicating its potential to generate descriptive captions of scenes.

\noindent\textbf{Task decomposition.}
Task decomposition involves breaking down a task into subtasks based on the 3D scene's spatial relationships. 
\systemname shows competitive performance in zero-shot style (Table~\ref{tab:held-in}), underscoring its ability to leverage spatial information effectively.

\noindent\textbf{3D semantic segmentation.}
3D semantic segmentation, a cornerstone for understanding complex environments, involves assigning semantic labels to each 3D point in a scene. Our investigation employs VLMs for zero-shot segmentation, showcasing the potential of VLMs to navigate this task without extensive labeled data.

Our approach yielded a mIoU of 5.6 using randomly selected images. By iteratively refining viewpoint selection, we improved the mIoU to 8.7, highlighting the importance of strategic view selection in enhancing zero-shot segmentation performance.

While our zero-shot method does not yet rival the accuracy of traditional 3D semantic segmentation techniques, it illustrates the untapped potential of leveraging VLMs for complex 3D perception challenges. This work not only demonstrates the adaptability of language models to perception tasks but also opens new research avenues in integrating linguistic and visual understanding.

\subsection{Qualitative Results}
\label{subsection:qualitative}

This section showcases two qualitative examples that highlight the efficacy and versatility of the proposed \lsystemname in addressing complex scenarios.

\noindent\textbf{Qualitative result for 3D reasoning.} 
Our first case study focuses on ScanQA, as illustrated in Figure~\ref{fig:cap}. This example vividly demonstrates \lsystemname's ability to accurately identify and describe objects and their relationships within 3D environments. Through intelligent scene analysis based on selected images, \lsystemname excels in synthesizing information from multiple objects to provide precise 3D scene captions, highlighting its advanced 3D reasoning capabilities. Additionally, \lsystemname can decompose tasks in specific environments with minimal supplementary information. Further details and results are available in the supplementary materials.

\noindent\textbf{Qualitative result for real-world navigation. }
A compelling application of \lsystemname is its utility in real-world navigation tasks. In this case study, we explore a scenario involving the navigation towards a printer located in a typical office setting. 
Uniquely, for this experiment, we forgo the initial BEV image input, challenging \lsystemname to iteratively engage with its surroundings to determine optimal viewpoints for progression. Although the office overview (depicted in the upper part of Figure~\ref{fig:demo}) is not directly fed into \lsystemname, it serves as a context for understanding the complexity of the navigation task, with the printer situated in the southeast direction amidst numerous obstacles.

\lsystemname adeptly navigates this environment by making informed decisions at each juncture, incorporating historical data to look around and select the most promising path forward. This process exemplifies the system's capacity to not only \textit{circumnavigate obstacles} efficiently but also to successfully identify and reach the target destination, all while recognizing essential objects like the printer. The VLM thus demonstrates a profound ability to explore and interact with an open-set world, leveraging solely image-based observations to understand and act within complex 3D scenes, as shown in Figure~\ref{fig:demo}.

\subsection{Ablation Study}
\label{subsection:ablation}

\begin{table*}[t]
\caption{\textbf{Effect of the number of viewpoints on 3D QA.} This experiment is evaluated on the validation set of ScanQA dataset.
% With more images, our method can understand 3D scenes more accurately. The number of images is limited by the ability of VLMs.
}
\vspace{-5pt}
    \centering
    % \small
    \setlength{\tabcolsep}{3.6mm}
    {
    \begin{tabular}{c|c|c|c|c|c}
        \hline
        \# Viewpoints & BLEU-1& METEOR & ROUHE-L & CIDEr & EM \\ \hline
        6 & 17.1 & 12.8 & 28.4 & 50.8 & 13.1 \\ 
        12 & 23.3 & 15.0 & 35.3 & 67.9 & 16.9 \\ 
        24 & \textbf{34.1} & \textbf{16.5} &  \textbf{40.3} & \textbf{82.0} & \textbf{21.1} \\ 
        \hline
    \end{tabular}  
    }
    \label{tab:ablation_num}
    % \vspace{-1em}
\end{table*}

\begin{table}[]
\caption{\textbf{Effect of the density of lines in Set-of-Line prompt on 3D QA.} This experiment is evaluated on the validation set of ScanQA dataset. Here, ``-'' indicates the GPT-4V fails to output the results.
    % `0x0' indicates no prompts, resulting in ineffective 3D scene interpretation by the VLM. Increasing line density improves understanding up to a limit, beyond which sticks obscure each other (16x16), diminishing effectiveness.
    }
    \vspace{-5pt}
    \centering
    % \small
    \setlength{\tabcolsep}{3.8mm}
    {
    \begin{tabular}{c|c|c|c|c|c}
        \hline
        Line density &  BLEU-1 &  METEOR & ROUGE-L & CIDEr & EM \\ \hline
        0x0 &  - &- & - & - & - \\ 
        4x4 &  23.2 &14.2 & 34.0 & 66.0 & 16.9 \\ 
        8x8 &  \textbf{34.1} & \textbf{16.5} & \textbf{40.3} & \textbf{82.0} & \textbf{21.1} \\ 
        16x16 &  - & - & - & - & - \\ 
        \hline
    \end{tabular}
    }
    \label{tab:ablation_density}
    \vspace{-5pt}
\end{table}

In this section, we present an ablation study designed to evaluate the influence of varying grid line prompts on the selection of camera poses, subsequently affecting the VLMs' capacity for 3D scene understanding. Specifically, we randomly select a subset comprising 20\% of the scenes from the ScanQA validation dataset as the basis for our investigation, focusing on the 3D question answering task.

\noindent\textbf{Effect of viewpoints number.} Our investigation into the influence of observation views on the performance of our system, herein referred to as \lsystemname, is meticulously documented in Table~\ref{tab:ablation_num}. This analysis prompts the VLM to interpret 3D scenes using selections of 6, 12, and 24 images, with the choice in image count being constrained by the computational capabilities of the VLM. Our findings indicate a direct correlation between the number of images and the performance outcome, where an increase in image count leads to notably enhanced results. This trend underscores the potential of augmenting input imagery to significantly improve the efficacy of our proposed approach.

\noindent\textbf{Effect of line density.} We further investigate the role of line density within the Set-of-Line prompts utilized by \lsystemname, with results detailed in Table~\ref{tab:ablation_density}. The application of line density is subject to the recognition abilities of the VLMs, imposing a practical limit to the density achievable (16x16). Despite this constraint, our results demonstrate that a higher density of auxiliary lines correlates with more precise 3D scene comprehension. The absence of dense visual prompts (0x0) significantly hampers the VLMs' ability to accurately determine camera poses, highlighting a notable limitation in the current capabilities of \lsystemname for precise and mathematical pose estimation.

\section{Conclusion}

In this work, we introduce \lsystemname, a pioneering framework that utilizes Vision-Language Models for zero-shot understanding and interaction within 3D environments. Through the strategic selection of diverse observational viewpoints and the incorporation of custom-designed visual prompts, \lsystemname facilitates a nuanced and integrated perception of 3D scenes. Our experiments underscore the transformative potential of VLMs in redefining 3D scene analysis, emphasizing the effectiveness of multi-viewpoint synthesis and visual prompts in augmenting model capabilities. This advancement propels us towards the realization of intelligent systems proficient in comprehending and navigating the real world akin to human interaction.

% \noindent\textbf{Limitation }

%
\clearpage
\bibliographystyle{splncs04}
\bibliography{main}

\begin{thebibliography}{10}
\providecommand{\url}[1]{\texttt{#1}}
\providecommand{\urlprefix}{URL }
\providecommand{\doi}[1]{https://doi.org/#1}

\bibitem{alayrac2022flamingo}
Alayrac, J.B., Donahue, J., Luc, P., Miech, A., Barr, I., Hasson, Y., Lenc, K., Mensch, A., Millican, K., Reynolds, M., et~al.: Flamingo: a visual language model for few-shot learning. Advances in Neural Information Processing Systems  \textbf{35},  23716--23736 (2022)

\bibitem{antol2015vqa}
Antol, S., Agrawal, A., Lu, J., Mitchell, M., Batra, D., Zitnick, C.L., Parikh, D.: Vqa: Visual question answering. In: Proceedings of the IEEE international conference on computer vision. pp. 2425--2433 (2015)

\bibitem{azuma2022scanqa}
Azuma, D., Miyanishi, T., Kurita, S., Kawanabe, M.: Scanqa: 3d question answering for spatial scene understanding. In: proceedings of the IEEE/CVF conference on computer vision and pattern recognition. pp. 19129--19139 (2022)

\bibitem{banerjee2005meteor}
Banerjee, S., Lavie, A.: Meteor: An automatic metric for mt evaluation with improved correlation with human judgments. In: Proceedings of the acl workshop on intrinsic and extrinsic evaluation measures for machine translation and/or summarization. pp. 65--72 (2005)

\bibitem{cadena2016robotic}
Cadena, C., Dick, A.R., Reid, I.D.: Multi-modal auto-encoders as joint estimators for robotics scene understanding. In: Robotics: Science and systems. vol.~5 (2016)

\bibitem{caesar2020nuscenes}
Caesar, H., Bankiti, V., Lang, A.H., Vora, S., Liong, V.E., Xu, Q., Krishnan, A., Pan, Y., Baldan, G., Beijbom, O.: nuscenes: A multimodal dataset for autonomous driving. In: Proceedings of the IEEE/CVF conference on computer vision and pattern recognition. pp. 11621--11631 (2020)

\bibitem{chen2024spatialvlm}
Chen, B., Xu, Z., Kirmani, S., Ichter, B., Driess, D., Florence, P., Sadigh, D., Guibas, L., Xia, F.: Spatialvlm: Endowing vision-language models with spatial reasoning capabilities. arXiv preprint arXiv:2401.12168  (2024)

\bibitem{chen2020scanrefer}
Chen, D.Z., Chang, A.X., Nie{\ss}ner, M.: Scanrefer: 3d object localization in rgb-d scans using natural language. In: European conference on computer vision. pp. 202--221. Springer (2020)

\bibitem{chen2023videollm}
Chen, G., Zheng, Y.D., Wang, J., Xu, J., Huang, Y., Pan, J., Wang, Y., Wang, Y., Qiao, Y., Lu, T., et~al.: Videollm: Modeling video sequence with large language models. arXiv preprint arXiv:2305.13292  (2023)

\bibitem{chen2023minigpt}
Chen, J., Zhu, D., Shen, X., Li, X., Liu, Z., Zhang, P., Krishnamoorthi, R., Chandra, V., Xiong, Y., Elhoseiny, M.: Minigpt-v2: large language model as a unified interface for vision-language multi-task learning. arXiv preprint arXiv:2310.09478  (2023)

\bibitem{chowdhery2023palm}
Chowdhery, A., Narang, S., Devlin, J., Bosma, M., Mishra, G., Roberts, A., Barham, P., Chung, H.W., Sutton, C., Gehrmann, S., et~al.: Palm: Scaling language modeling with pathways. Journal of Machine Learning Research  \textbf{24}(240),  1--113 (2023)

\bibitem{chung2022flant5}
Chung, H.W., Hou, L., Longpre, S., Zoph, B., Tay, Y., Fedus, W., Li, Y., Wang, X., Dehghani, M., Brahma, S., et~al.: Scaling instruction-finetuned language models. arXiv preprint arXiv:2210.11416  (2022)

\bibitem{curless1996reconstruct}
Curless, B., Levoy, M.: A volumetric method for building complex models from range images. In: Proceedings of the 23rd annual conference on Computer graphics and interactive techniques. pp. 303--312 (1996)

\bibitem{dai2017scannet}
Dai, A., Chang, A.X., Savva, M., Halber, M., Funkhouser, T., Nie{\ss}ner, M.: Scannet: Richly-annotated 3d reconstructions of indoor scenes. In: Proceedings of the IEEE conference on computer vision and pattern recognition. pp. 5828--5839 (2017)

\bibitem{deitke2023objaverse}
Deitke, M., Schwenk, D., Salvador, J., Weihs, L., Michel, O., VanderBilt, E., Schmidt, L., Ehsani, K., Kembhavi, A., Farhadi, A.: Objaverse: A universe of annotated 3d objects. In: Proceedings of the IEEE/CVF Conference on Computer Vision and Pattern Recognition. pp. 13142--13153 (2023)

\bibitem{ding2019votenet}
Ding, Z., Han, X., Niethammer, M.: Votenet: A deep learning label fusion method for multi-atlas segmentation. In: Medical Image Computing and Computer Assisted Intervention--MICCAI 2019: 22nd International Conference, Shenzhen, China, October 13--17, 2019, Proceedings, Part III 22. pp. 202--210. Springer (2019)

\bibitem{geiger2015kitti}
Geiger, A., Lenz, P., Stiller, C., Urtasun, R.: The kitti vision benchmark suite. URL http://www. cvlibs. net/datasets/kitti  \textbf{2}(5) (2015)

\bibitem{gong2023multimodalgpt}
Gong, T., Lyu, C., Zhang, S., Wang, Y., Zheng, M., Zhao, Q., Liu, K., Zhang, W., Luo, P., Chen, K.: Multimodal-gpt: A vision and language model for dialogue with humans. arXiv preprint arXiv:2305.04790  (2023)

\bibitem{guo2021atsurvey}
Guo, Z., Huang, Y., Hu, X., Wei, H., Zhao, B.: A survey on deep learning based approaches for scene understanding in autonomous driving. Electronics  \textbf{10}(4), ~471 (2021)

\bibitem{hong20243dllm}
Hong, Y., Zhen, H., Chen, P., Zheng, S., Du, Y., Chen, Z., Gan, C.: 3d-llm: Injecting the 3d world into large language models. Advances in Neural Information Processing Systems  \textbf{36} (2024)

\bibitem{horaud2016depthcamera}
Horaud, R., Hansard, M., Evangelidis, G., M{\'e}nier, C.: An overview of depth cameras and range scanners based on time-of-flight technologies. Machine vision and applications  \textbf{27}(7),  1005--1020 (2016)

\bibitem{hua2017objectNN}
Hua, B.S., Truong, Q.T., Tran, M.K., Pham, Q.H., Kanezaki, A., Lee, T., Chiang, H., Hsu, W., Li, B., Lu, Y., et~al.: Shrec’17: Rgb-d to cad retrieval with objectnn dataset. In: Proc. Eurograph. Workshop 3D Object Retrieval. pp. 25--32 (2017)

\bibitem{jatavallabhula2023conceptfusion}
Jatavallabhula, K.M., Kuwajerwala, A., Gu, Q., Omama, M., Chen, T., Maalouf, A., Li, S., Iyer, G., Saryazdi, S., Keetha, N., et~al.: Conceptfusion: Open-set multimodal 3d mapping. arXiv preprint arXiv:2302.07241  (2023)

\bibitem{kalkofen2008ar}
Kalkofen, D., Mendez, E., Schmalstieg, D.: Comprehensible visualization for augmented reality. IEEE transactions on visualization and computer graphics  \textbf{15}(2),  193--204 (2008)

\bibitem{kim2021vilt}
Kim, W., Son, B., Kim, I.: Vilt: Vision-and-language transformer without convolution or region supervision (2021)

\bibitem{lemmens2007lidar}
Lemmens, M.: Airborne lidar sensors. GIM international  \textbf{21}(2),  24--27 (2007)

\bibitem{li2022mplug}
Li, C., Xu, H., Tian, J., Wang, W., Yan, M., Bi, B., Ye, J., Chen, H., Xu, G., Cao, Z., et~al.: mplug: Effective and efficient vision-language learning by cross-modal skip-connections. arXiv preprint arXiv:2205.12005  (2022)

\bibitem{li2023blip2}
Li, J., Li, D., Savarese, S., Hoi, S.: Blip-2: Bootstrapping language-image pre-training with frozen image encoders and large language models. arXiv preprint arXiv:2301.12597  (2023)

\bibitem{li2022blip}
Li, J., Li, D., Xiong, C., Hoi, S.: Blip: Bootstrapping language-image pre-training for unified vision-language understanding and generation. In: International Conference on Machine Learning. pp. 12888--12900. PMLR (2022)

\bibitem{li20243dmit}
Li, Z., Zhang, C., Wang, X., Ren, R., Xu, Y., Ma, R., Liu, X.: 3dmit: 3d multi-modal instruction tuning for scene understanding. arXiv preprint arXiv:2401.03201  (2024)

\bibitem{lin2004rouge}
Lin, C.Y.: Rouge: A package for automatic evaluation of summaries. In: Text summarization branches out. pp. 74--81 (2004)

\bibitem{liu2024llava}
Liu, H., Li, C., Wu, Q., Lee, Y.J.: Visual instruction tuning. Advances in neural information processing systems  \textbf{36} (2024)

\bibitem{naeem2023i2mvformer}
Naeem, M.F., Khan, M.G.Z.A., Xian, Y., Afzal, M.Z., Stricker, D., Van~Gool, L., Tombari, F.: I2mvformer: Large language model generated multi-view document supervision for zero-shot image classification. In: Proceedings of the IEEE/CVF Conference on Computer Vision and Pattern Recognition. pp. 15169--15179 (2023)

\bibitem{naseer2018survey}
Naseer, M., Khan, S., Porikli, F.: Indoor scene understanding in 2.5/3d for autonomous agents: A survey. IEEE access  \textbf{7},  1859--1887 (2018)

\bibitem{openai2023gpt4}
OpenAI: Gpt-4 technical report (2023)

\bibitem{ouyang2022InstructGPT}
Ouyang, L., Wu, J., Jiang, X., Almeida, D., Wainwright, C., Mishkin, P., Zhang, C., Agarwal, S., Slama, K., Ray, A., et~al.: Training language models to follow instructions with human feedback. Advances in Neural Information Processing Systems  \textbf{35},  27730--27744 (2022)

\bibitem{papineni2002bleu}
Papineni, K., Roukos, S., Ward, T., Zhu, W.J.: Bleu: a method for automatic evaluation of machine translation. In: Proceedings of the 40th annual meeting of the Association for Computational Linguistics. pp. 311--318 (2002)

\bibitem{radford2021clip}
Radford, A., Kim, J.W., Hallacy, C., Ramesh, A., Goh, G., Agarwal, S., Sastry, G., Askell, A., Mishkin, P., Clark, J., et~al.: Learning transferable visual models from natural language supervision. In: International conference on machine learning. pp. 8748--8763. PMLR (2021)

\bibitem{ramakrishnan2021habitat}
Ramakrishnan, S.K., Gokaslan, A., Wijmans, E., Maksymets, O., Clegg, A., Turner, J., Undersander, E., Galuba, W., Westbury, A., Chang, A.X., et~al.: Habitat-matterport 3d dataset (hm3d): 1000 large-scale 3d environments for embodied ai. arXiv preprint arXiv:2109.08238  (2021)

\bibitem{rasheed2023glamm}
Rasheed, H., Maaz, M., Shaji, S., Shaker, A., Khan, S., Cholakkal, H., Anwer, R.M., Xing, E., Yang, M.H., Khan, F.S.: Glamm: Pixel grounding large multimodal model. arXiv preprint arXiv:2311.03356  (2023)

\bibitem{schwenk2022okvqa}
Schwenk, D., Khandelwal, A., Clark, C., Marino, K., Mottaghi, R.: A-okvqa: A benchmark for visual question answering using world knowledge. In: European Conference on Computer Vision. pp. 146--162. Springer (2022)

\bibitem{vedantam2015cider}
Vedantam, R., Lawrence~Zitnick, C., Parikh, D.: Cider: Consensus-based image description evaluation. In: Proceedings of the IEEE conference on computer vision and pattern recognition. pp. 4566--4575 (2015)

\bibitem{wang2023chat}
Wang, Z., Huang, H., Zhao, Y., Zhang, Z., Zhao, Z.: Chat-3d: Data-efficiently tuning large language model for universal dialogue of 3d scenes. arXiv preprint arXiv:2308.08769  (2023)

\bibitem{wu2023gpt4vis}
Wu, W., Yao, H., Zhang, M., Song, Y., Ouyang, W., Wang, J.: Gpt4vis: What can gpt-4 do for zero-shot visual recognition? arXiv preprint arXiv:2311.15732  (2023)

\bibitem{wu2015modelnet}
Wu, Z., Song, S., Khosla, A., Yu, F., Zhang, L., Tang, X., Xiao, J.: 3d shapenets: A deep representation for volumetric shapes. In: Proceedings of the IEEE conference on computer vision and pattern recognition. pp. 1912--1920 (2015)

\bibitem{yang2023som}
Yang, J., Zhang, H., Li, F., Zou, X., Li, C., Gao, J.: Set-of-mark prompting unleashes extraordinary visual grounding in gpt-4v. arXiv preprint arXiv:2310.11441  (2023)

\bibitem{yang2023set}
Yang, J., Zhang, H., Li, F., Zou, X., Li, C., Gao, J.: Set-of-mark prompting unleashes extraordinary visual grounding in gpt-4v. arXiv preprint arXiv:2310.11441  (2023)

\bibitem{yu2019mcan}
Yu, Z., Yu, J., Cui, Y., Tao, D., Tian, Q.: Deep modular co-attention networks for visual question answering. In: Proceedings of the IEEE/CVF conference on computer vision and pattern recognition. pp. 6281--6290 (2019)

\bibitem{zhang2023llama}
Zhang, R., Han, J., Zhou, A., Hu, X., Yan, S., Lu, P., Li, H., Gao, P., Qiao, Y.: Llama-adapter: Efficient fine-tuning of language models with zero-init attention. arXiv preprint arXiv:2303.16199  (2023)

\bibitem{zhang2024hvdistill}
Zhang, S., Deng, J., Bai, L., Li, H., Ouyang, W., Zhang, Y.: Hvdistill: Transferring knowledge from images to point clouds via unsupervised hybrid-view distillation. International Journal of Computer Vision pp. 1--15 (2024)

\end{thebibliography}

\end{document}